\documentclass[conference]{IEEEconf}
\IEEEoverridecommandlockouts
\usepackage[utf8]{inputenc}

\usepackage{cite}
\usepackage{amsmath,amssymb,amsfonts}

\usepackage{graphicx}
\usepackage{textcomp}
\usepackage{xcolor}
\usepackage{bm}
\usepackage{multirow}
\usepackage{float}
\usepackage{tikz}
\usepackage{booktabs}
\makeatletter
\let\NAT@parse\undefined
\def\bstctlcite{\@ifnextchar[{\@bstctlcite}{\@bstctlcite[@auxout]}}
\def\@bstctlcite[#1]#2{\@bsphack
  \@for\@citeb:=#2\do{%
    \edef\@citeb{\expandafter\@firstofone\@citeb}%
    \if@filesw\immediate\write\csname #1\endcsname{\string\citation{\@citeb}}\fi}%
  \@esphack}
\let\IEEEoriginalmakecaption\@makecaption
\long\def\@makecaption#1#2{{%
  \let\footnotesize\normalsize
  \IEEEoriginalmakecaption{#1}{#2}%
}}
\makeatother
\def\BibTeX{{\rm B\kern-.05em{\sc i\kern-.025em b}\kern-.08em
    T\kern-.1667em\lower.7ex\hbox{E}\kern-.125emX}}

\usepackage[
    colorlinks=true,
    linkcolor=blue,
    citecolor=blue,
    urlcolor=blue
]{hyperref}
\begin{document}
\bstctlcite{IEEEtran:BSTcontrol}

\definecolor{C1}{HTML}{00A6FF}
\definecolor{C2}{HTML}{1E9BF0}
\definecolor{C3}{HTML}{4F8FE8}
\definecolor{C4}{HTML}{7279E0}
\definecolor{C5}{HTML}{8B63D9}
\definecolor{C6}{HTML}{B85BC0}
\definecolor{C7}{HTML}{E0648D}
\definecolor{C8}{HTML}{FF7A45}

\newcommand{\TouchTherm}{%
  \textbf{%
    \textcolor{C1}{T}%
    \textcolor{C2}{o}%
    \textcolor{C3}{u}%
    \textcolor{C4}{c}%
    \textcolor{C5}{h}%
    \textcolor{C6}{T}%
    \textcolor{C7}{h}%
    \textcolor{C8}{e}%
    \textcolor{C8}{r}%
    \textcolor{C8}{m}%
  }%
}

\title{\TouchTherm: Building Multimodal Digital Twins of Objects for Tactile and Thermal Rendering}

\author{
Yitao Zhang, Hong Ying, Haoran Guo, Xiaoying Zhou, Guanyu Chen, and Chenxi Xiao$^\dagger$%
\thanks{$^\dagger$ Corresponding author.}%
\thanks{This work was supported by the Natural Science Foundation of Shanghai
(Grant No. 25ZR1402370), and partially by Shanghai Frontiers Science
Center of Human-centered Artificial Intelligence (ShangHAI), MoE Key
Laboratory of Intelligent Perception and Human-Machine Collaboration
(KLIP-HuMaCo).}%
\thanks{Yitao Zhang, Hong Ying, Haoran Guo, Xiaoying Zhou, Guanyu Chen, and Chenxi Xiao are with the School of Information Science and Technology, ShanghaiTech University, Shanghai 201210, China.  \texttt{\{zhangyt2025, yinghong2023, guohr2026, zhouxy2023, chengy2025, xiaochx\}@shanghaitech.edu.cn}.}%
\thanks{Project webpage: \href{https://anonymous-research1.github.io/}{https://anonymous-research1.github.io/}}%
}

\maketitle

\begin{abstract}
Robotic simulation and virtual reality increasingly require object assets that capture not only visual geometry, but also the physical cues underlying tactile and thermal interaction. However, existing 3D datasets and reconstruction methods primarily represent object-scale geometry and visual appearance, overlooking the microscale surface structure needed for high-fidelity haptic rendering and the transient temperature dynamics required for temperature-aware interaction. We present \textbf{TouchTherm}, a framework for constructing simulation-ready visuo-tactile-thermal object assets from real-world objects. For visual and tactile reconstruction, we combine a high-fidelity handheld structured-light scanner with multiview normal maps obtained from photometric stereo. The normal maps are registered to the scanned geometry and transformed into tangent space to recover local micro-height fields, which are queried by an optical tactile renderer, while the coarse mesh handles collision detection. For thermal reconstruction, we capture synchronized multiview infrared videos of natural cooling following controlled heating and reconstruct a physics-regularized dynamic thermal field. These representations are integrated with the visual geometry into a unified, simulator-ready object asset. Experiments on 20 objects show that the reconstructed micro-height fields preserve dominant surface structures and recover higher-frequency details beyond the coarse geometry, while the dynamic thermal fields achieve held-out surface-temperature MAEs of $0.465~^{\circ}\mathrm{C}$ and $0.592~^{\circ}\mathrm{C}$ at 30~s and 45~s, respectively. The resulting tactile assets enable high-fidelity tactile simulation, including synthetic-to-real object recognition from tactile sensor observations, while a glove-based VR system demonstrates spatially and temporally varying thermal feedback. These results highlight the potential of \textbf{TouchTherm} for multimodal sensory simulation and temperature-aware virtual interaction.
\end{abstract}

\section{Introduction}

Simulation has become a foundational tool for robot learning, but its effectiveness remains limited by the reality gap in contact-rich interactions~\cite{makoviychuk2021isaacgym,zhao2020simtoreal}. One contributor to this gap is the limited range of sensory cues that virtual objects can provide during interaction. Existing digital assets generally supply object-scale geometry for visual rendering and force-based contact simulation, but often omit object-side representations of contact-scale tactile appearance and transient surface temperature~\cite{si2021taxim,si2024difftactilephysicsbaseddifferentiabletactile}. These cues are relevant to robotic perception, human--robot interaction, virtual reality, and teleoperation~\cite{yuan2017gelsight}. We therefore seek to augment conventional geometric assets with spatially aligned tactile-rendering and thermal-feedback attributes. However, constructing such assets presents two primary challenges:

\begin{figure}[!t]
    \centering
    \includegraphics[width=\linewidth]{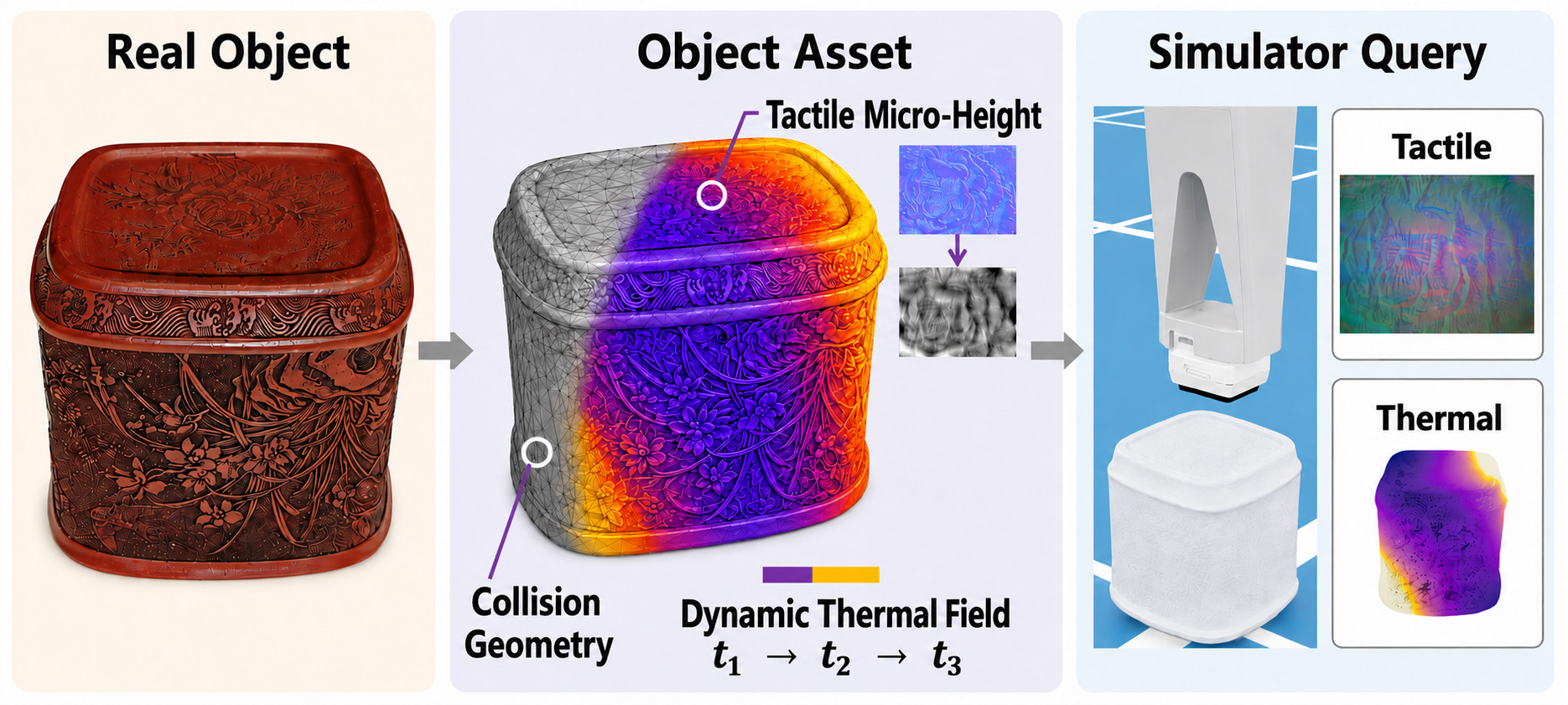}
    \caption{\textbf{Overview of TouchTherm}. From real-object observations, our pipeline constructs simulation-ready assets combining visual geometry, contact-aligned tactile microgeometry, and observation-driven dynamic thermal fields for optical tactile rendering and temperature-field simulation.}
    \label{fig:teaser}
    \vspace{-1.4\baselineskip}
\end{figure}

\textbf{(1) Limitations of microgeometry acquisition and geometry representation tradeoffs.} 
Existing acquisition methods provide accurate object-scale geometry and visual appearance~\cite{schoenberger2016sfm,mildenhall2020nerf,kerbl2023gaussian}, but contact-relevant micro-structures such as shallow ridges, embossing, grooves, and anisotropic textures can be attenuated by finite scanning resolution and reconstruction artifacts despite their strong influence on optical tactile measurements~\cite{yuan2017gelsight,taylor2022gelslim}. Encoding such micro-relief directly as dense polygonal geometry would also increase storage and simulation costs~\cite{chen2023tacchi}. We therefore seek to preserve contact-scale surface relief separately from object-scale collision geometry.

\textbf{(2) Lack of object-level dynamic thermal field representations.}
Thermal cues contribute to material perception and can enhance thermal interaction in virtual and remote environments~\cite{watkins2026immersivewearablethermalrendering,Bha2021toh}. However, conventional visual assets do not encode how surface temperature varies across an object or evolves during heating and cooling~\cite{abad2021haptitemp,Par2021afm}. These dynamics depend on object-specific material properties, geometry, and therefore cannot be represented only by a single temperature value or material label. A useful multimodal object asset should instead provide a spatially varying dynamic thermal field that supports surface-temperature queries over time.

To bridge these gaps, we introduce \textbf{TouchTherm}, a pipeline for constructing simulation-ready visuo-tactile-thermal object assets from real objects (Fig.~\ref{fig:teaser}). Visual object datasets primarily provide geometry and appearance~\cite{calli2015ycb,downs2022google}, whereas multisensory datasets typically contain partial, observation-only data~\cite{gao2021objectfolder,gao2022objectfolder2}. In contrast, TouchTherm reconstructs spatially varying tactile fields and dynamic thermal fields at the object level. For tactile rendering, a coarse collision mesh is augmented with local tangent-space micro-height fields, preserving contact-scale relief without dense collision geometry. For thermal simulation, the reconstructed dynamic thermal field supports time-varying surface-temperature queries for robotic and virtual interactions.

In summary, our main contributions are as follows. 
\begin{itemize}
\item \textbf{Multimodal object twins:} We introduce a pipeline for constructing simulation-ready visuo-tactile-thermal object assets with registered visual geometry, tactile micro-height fields, and dynamic thermal fields.

\item \textbf{Tactile rendering:} We develop a contact-conditioned pipeline for rendering optical tactile observations from reconstructed micro-height fields.

\item \textbf{Dynamic thermal field rendering:} We reconstruct the field from multiview infrared cooling videos and identify effective thermal parameters for time-varying surface-temperature queries.

\item \textbf{Validation and application:} We validate the framework on 20 objects through tactile appearance, thermal prediction, and synthetic-to-real recognition, and demonstrate wearable thermal feedback in VR.
\end{itemize}

\section{Related Work}
\subsection{Multimodal Data Collection}

Existing object reconstruction pipelines are largely vision-centric. Structured-light scanning and multi-view stereo recover metric geometry~\cite{schoenberger2016sfm}, while NeRF and 3D Gaussian Splatting enable view synthesis and high-fidelity rendering~\cite{mildenhall2020nerf,kerbl2023gaussian}. However, these methods mainly capture object-scale geometry and appearance, overlooking the contact-scale surface relief and transient thermal properties needed for multimodal sensory rendering.

Non-visual sensing provides complementary physical information. Tactile sensors capture local surface geometry and contact responses~\cite{yuan2017gelsight,taylor2022gelslim,lin2023dtact}, while thermal sensors capture spatial temperature distributions and heat-transfer dynamics~\cite{abad2021haptitemp}. Yet, these measurements are typically localized observations rather than object-level representations registered to the object geometry. We therefore construct a unified representation that spatially registers visual, tactile, and thermal information as reusable simulation assets.

\subsection{Datasets and Simulation Assets}

\textbf{Geometry and appearance assets.}
Existing robotics and embodied-AI datasets primarily provide object-scale geometry and visual appearance. YCB provides standardized physical objects with RGB-D scans~\cite{calli2015ycb}, Google Scanned Objects contains reconstructed household objects suitable for physics simulation~\cite{downs2022google}, and large-scale collections such as Objaverse and Digital Twin Catalog provide extensive 3D assets for learning and simulation
~\cite{deitke2023objaverse,dong2025digital}. These resources substantially expand the diversity of available digital objects, but generally lack spatially registered contact-scale tactile relief and dynamic surface-temperature information.

\textbf{Multisensory object assets.}
ObjectFolder and its extensions associate objects with visual, acoustic, and tactile observations or implicit representations~\cite{gao2021objectfolder,gao2022objectfolder2,gao2023objectfolderbenchmark}, while X-Capture collects synchronized RGB-D, tactile, and impact-audio measurements of physical objects~\cite{clarke2025xcapture}. These datasets primarily support learning from existing multisensory observations. However, none combines registered tactile microgeometry with spatiotemporal thermal responses in a unified object asset.

\textbf{Simulation-ready tactile and thermal assets.}
Tactile simulators synthesize contact observations from object geometry or height-field representations~\cite{si2021taxim,chen2023tacchi,gomes2023beyond,si2024difftactilephysicsbaseddifferentiabletactile}. Coarse meshes are efficient for collision and contact simulation but may miss tactile-relevant surface details~\cite{geng2011structured,schoenberger2016sfm}, while dense geometric representations of microscale structures can be used for haptic rendering but increase simulation and rendering costs~\cite{chen2023tacchi,si2024difftactilephysicsbaseddifferentiabletactile}. Beyond tactile-texture simulation, thermal simulation is less developed: prior work mainly uses thermal cues for material recognition or user feedback, rather than constructing object-level dynamic thermal fields. \textbf{TouchTherm} bridges these gaps by combining coarse collision geometry, registered tactile micro-height fields, and an object-specific dynamic thermal field within a shared object coordinate frame.

\begin{figure*}[t]
    \centering
    \includegraphics[width=\textwidth]{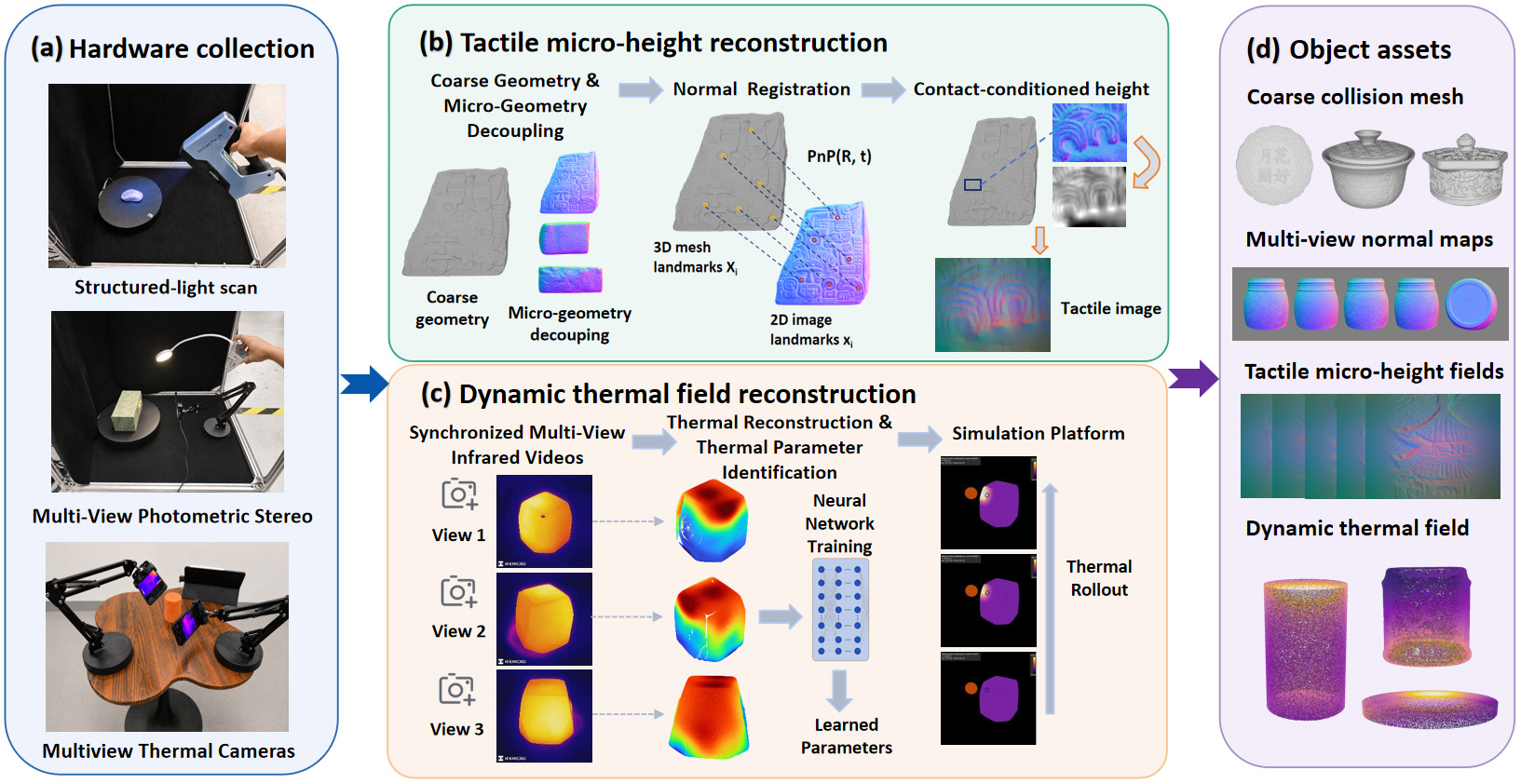}
    \caption{\textbf{Overview of the TouchTherm reconstruction pipeline.} (a) Structured-light scanning, multiview photometric stereo, and thermal imaging acquire geometry, normals, and infrared videos. (b) Coarse geometry is decoupled from registered microgeometry, reconstructed into contact-conditioned height fields for tactile rendering. (c) Infrared observations support physics-regularized dynamic thermal reconstruction and parameter identification for rollout. (d) The asset comprises a coarse collision mesh, normal maps, tactile micro-height fields, and a dynamic thermal field.}
    \label{fig:method_overview}
    \vspace{-1.0\baselineskip}
\end{figure*}

\section{Methodology}

\subsection{Pipeline Overview}
\label{sec:pipeline_overview}

Our goal is to convert a real object into a simulator-ready multimodal asset comprising a coarse mesh, tactile micro-height fields, and a dynamic thermal field. As shown in Fig.~\ref{fig:method_overview}, the pipeline aligns object scans, multiview normal maps, and infrared videos to a common object frame, then reconstructs tactile microgeometry and the dynamic thermal field in two parallel branches. The resulting representations are integrated into a unified object asset. According to this pipeline, the following sections describe the acquisition setup (Sec.~\ref{sec:hardware_setup}), tactile reconstruction (Sec.~\ref{sec:tactile_reconstruction}), and dynamic thermal field reconstruction (Sec.~\ref{sec:thermal_reconstruction}), respectively.

\subsection{Hardware Setup}
\label{sec:hardware_setup}
As shown in Fig.~\ref{fig:method_overview}(a), we use an EinScan Pro 2X V2 structured-light scanner to acquire an
object mesh. For appearance and normal estimation, a fixed Redmi K70
smartphone captures a stack of RGB images at each viewpoint under manually
varied illumination directions. The pretrained SDM-UniPS~\cite{ikehata2023sdmunips}
model then estimates an image-aligned normal map from each illumination
stack. For thermal acquisition, HIKMICRO P09 thermal cameras
record the natural cooling process following controlled heat-gun excitation.
The RGB and thermal cameras are calibrated, and registered to the object
frame using manually selected 2D--3D
correspondences. Finally, a GelSight Mini provides tactile measurements for
validation and calibration of the micro-height amplitude of the reconstructed
surface.

\subsection{Reconstruction of Tactile Microgeometry for Rendering}
\label{sec:tactile_reconstruction}

As shown in Fig.~\ref{fig:method_overview}(b), we decouple contact geometry from the tactile
appearance on the scale of tactile sensor observations. Specifically, the coarse mesh provides the geometric foundation for
visual rendering and physical collision, while registered local micro-height
fields encode high-frequency surface relief exclusively for tactile rendering.

\subsubsection{Coarse Geometry and Microgeometry Decoupling}

The scanned mesh is first cleaned by removing support surfaces, isolated
fragments, and reconstruction artifacts.
For simple geometries, measured dimensions can alternatively be used to
construct a cleaner mesh. The resulting mesh
$\mathcal{M}=(V,F)$ is exported to Isaac Sim as the visual and collision
geometry. Next, contact-scale surface texture is represented separately as local
micro-height fields, thereby preserving high-frequency tactile detail
without increasing the geometric or collision complexity of the simulation
mesh. Detailed process for obtaining texture is provided as follows.

\subsubsection{Normal Estimation and Registration}
\label{sec:normal_estimation_registration}
For each RGB view $i$, the SDM-UniPS model \cite{ikehata2023sdmunips} estimates an image-aligned normal map
$\mathbf N_i$ from multi-illumination images. We manually
annotate sparse 2D--3D correspondences between the RGB \& normal image and the coarse
mesh, and estimate the object-to-camera pose $(\mathbf R_i,\mathbf t_i)$ using PnP. An
object-space point $\mathbf x$ is projected into view $i$ as
\begin{equation}
\mathbf q_i(\mathbf x)=\pi\!\left(\mathbf K_i(\mathbf R_i\mathbf x+\mathbf t_i)\right),
\end{equation}
where $\mathbf K_i$ denotes the camera intrinsic matrix and $\pi(\cdot)$ is the
perspective projection operator.

\subsubsection{Contact-conditioned Tactile Rendering}
\hfill\break
\hspace*{\parindent}\textbf{Metric surface-patch sampling.}
At a contact point $\mathbf p\in\mathcal{M}$, we define a local tangent frame
$(\mathbf t_u,\mathbf t_v,\mathbf n_0)$ and sample a metric surface patch
$\mathbf x(u,v)=\Psi_{\mathbf p}(u,v)$, where $(u,v)\in\Omega$ and
$\Omega=[-W/2,W/2]\times[-H/2,H/2]$. The patch dimensions $W$ and $H$
define the physical sensing area. The mapping $\Psi_{\mathbf p}$ is
implemented using closest-surface projection. Concurrently, we bilinearly sample the normal map at the projected location and transform the sampled normal into the local tangent frame:
\begin{equation}
\mathbf n^t(u,v)=
\begin{bmatrix}
\mathbf t_u^\top\\
\mathbf t_v^\top\\
\mathbf n_0^\top
\end{bmatrix}
\mathbf R_i^\top
\mathbf N_i\!\left(\mathbf q_i(\mathbf x(u,v))\right)
=
\begin{bmatrix}
n_u^t & n_v^t & n_w^t
\end{bmatrix}^{\top}.
\end{equation}

The local height gradients are given by
\begin{equation}
g_u=-\frac{n_u^t}{n_w^t+\epsilon}, \qquad
g_v=-\frac{n_v^t}{n_w^t+\epsilon},
\end{equation}
where $\epsilon$ is a small constant that prevents division by zero.

\textbf{Tangent-space height reconstruction.}
Due to errors in the estimated normal field, we recover the optimal
least-squares height field using Poisson reconstruction:
\begin{equation}
\begin{aligned}
h^\star
&=
\arg\min_h
\int_{\Omega}
\left\|
\nabla h-
\begin{bmatrix}
g_u\\
g_v
\end{bmatrix}
\right\|_2^2\,du\,dv,\\
h_\mu
&=
h^\star-\Pi(h^\star).
\end{aligned}
\end{equation}
where $\Pi(\cdot)$ extracts the low-frequency component of $h^\star$, leaving the micro-height residual $h_\mu$.  The least-squares problem is solved using a Poisson/DCT solver \cite{Sim1990TPAM}.

\textbf{Contact-conditioned composition.}
\label{sec:contact_tactile_rendering}
At rendering time, we composite the texture with the coarse geometry:
\begin{equation}
H_{\mathrm{render}}(u,v)
=
B(H_0,M)
+
\gamma M(u,v)h_\mu(u,v).
\label{eq:height_mix}
\end{equation}
Here, $B(H_0,M)$ denotes the resulting low-frequency indentation obtained from the coarse mesh $H_0(u,v)$ after applying the contact mask $M(u,v)\in[0,1]$. The dimensionless factor $\gamma$ is a Real2Sim calibration factor, selected once per object by aligning the rendered output with a GelSight observation from a separate calibration contact.

\begingroup

\subsection{Dynamic Thermal Field Reconstruction}
\label{sec:thermal_reconstruction}

We reconstruct the object's surface temperature field from multiview thermal videos and estimate two parameters: the surface-diffusion coefficient $\alpha$, which controls heat spreading along the surface, and the ambient-relaxation coefficient $h$, which controls heat exchange with the environment. These parameters drive the subsequent thermal simulation.

\textbf{1) Fusing thermal observations.}
We first map measurements from different cameras onto a common object surface. Each thermal camera is calibrated following~\cite{vidas2012mask}. Its pose relative to the object is estimated using RANSAC-PnP~\cite{fischler1981random}, then refined jointly with the other camera poses.

We correct pixel temperatures for temporal drift and viewing-angle bias following~\cite{lin2018thermal}, then project them onto the surface. At each surface point $\mathbf x_i$ and time $t$, we average the valid measurements from cameras observing that point to obtain $T_i^{\mathrm{obs}}(t)$. Points unseen by all cameras remain unobserved.

\textbf{2) Modeling surface diffusion.}
To model heat spreading along the surface, we define heat transfer between neighboring points. Let $\{\mathbf x_i\}_{i=1}^{N}$ be the object's point cloud, where $N$ is the number of points. We connect each point to its $k$ nearest neighbors, denoted by $\mathcal N_i$, and assign Gaussian weights:
\begin{equation}
w_{ij}
=
\exp\!\left(-\frac{\|\mathbf x_i-\mathbf x_j\|_2^2}{4t_g}\right).
\end{equation}
The weight $w_{ij}$ measures the connection strength between points $i$ and $j$ and decreases with distance. The parameter $t_g>0$ sets the spatial scale and has units of squared length. To reduce the influence of uneven point density, we normalize the weights as $\widetilde w_{ij}=w_{ij}/(\rho_i\rho_j)$, where $\rho_i=\sum_{j\in\mathcal N_i}w_{ij}$ is the local weight sum at point $i$.

We compute diffusion by comparing each point's temperature with those of its neighbors. Let $\mathbf T=[T_1,\ldots,T_N]^\top$ collect all surface temperatures. We define the graph diffusion operator $\mathbf L$ by
\begin{equation}
(\mathbf L\mathbf T)_i
=
\frac{
\sum_{j\in\mathcal N_i}\widetilde w_{ij}(T_i-T_j)
}{
t_g\sum_{j\in\mathcal N_i}\widetilde w_{ij}
}.
\end{equation}
Here, $(\mathbf L\mathbf T)_i$ is the difference between $T_i$ and the weighted average of its neighbors' temperatures, divided by $t_g$. It is positive when point $i$ is hotter than this average and negative when it is colder. Thus, the diffusion term $-\alpha\mathbf L\mathbf T$ cools the point in the first case and warms it in the second. The operator $\mathbf L$ approximates the negative surface Laplacian.

We also use $\mathbf L$ to construct spatial inputs for the temperature network. Following Delta-PINN~\cite{sahli2024delta}, we select its $K$ lowest-frequency eigenvectors. Each assigns a scalar value to every surface point and varies smoothly between neighbors. Collecting these $K$ values at $\mathbf x_i$ gives the feature vector $\boldsymbol\Phi(\mathbf x_i)\in\mathbb R^K$, which describes the point's location through the surface connectivity.

\textbf{3) Joint field reconstruction and parameter estimation.}
\label{sec:physics_reg}
To estimate temperatures in unobserved regions and identify $\alpha$ and $h$, we represent the temperature field with a multilayer perceptron (MLP):
\begin{equation}
\widehat T_i(t)
=
\mathcal N_\theta\!\left(\boldsymbol\Phi(\mathbf x_i),t\right),
\end{equation}
where $\theta$ denotes the network parameters. The network takes the spatial features and time as inputs and predicts the temperature $\widehat T_i(t)$ at point $i$. This representation lets us compute time derivatives from the learned field rather than directly from noisy measurements.
We jointly optimize $\theta$, $\alpha$, and $h$ using two losses. The data loss fits predicted temperatures to $T_i^{\mathrm{obs}}(t)$ wherever measurements are available. The physics loss encourages agreement with the thermal dynamics by penalizing the residual
\begin{equation}
r_i(t)
=
\frac{\partial\widehat T_i}{\partial t}
+\alpha(\mathbf L\widehat{\mathbf T})_i
+h(\widehat T_i-T_\infty),
\end{equation}
where $\widehat{\mathbf T}$ contains predictions at all surface points and $T_\infty$ is the ambient temperature. The residual is zero when the predicted temperature changes match those caused by surface diffusion and heat exchange with the environment.

\textbf{4) Runtime rollout.}
Once $\alpha$ and $h$ are identified, the heat-transfer model can advance the temperature field directly. We therefore discard the network and retain $\mathbf L$ and the estimated coefficients, with $\alpha$ in $\mathrm{m}^2/\mathrm{s}$ and $h$ in $\mathrm{s}^{-1}$.

Starting from an initial temperature vector $\mathbf T^0$, we update the field using explicit Euler integration:
\begin{equation}
\mathbf T^{n+1}
=
\mathbf T^n
-\Delta t\left[
\alpha\mathbf L\mathbf T^n
+h(\mathbf T^n-T_\infty\mathbf 1)
\right].
\end{equation}
Here, $\mathbf T^n$ contains the surface temperatures at step $n$, $\Delta t$ is the time step, and $\mathbf 1\in\mathbb R^N$ is the all-ones vector. Each update computes the next temperature field from the current field by accounting for surface diffusion and heat exchange with the environment.
 
\endgroup

\section{Experiments}
We evaluate TouchTherm from four perspectives: object asset quality
(Sec.~\ref{sec:exp_dataset}), tactile rendering fidelity
(Sec.~\ref{sec:exp_tactile_realism}), thermal prediction accuracy
(Sec.~\ref{sec:exp_thermal}), and downstream utility in perception and interaction
tasks (Sec.~\ref{sec:exp_tactile_recognition}).

\begin{figure}[t]
    \centering
    \includegraphics[width=\linewidth]{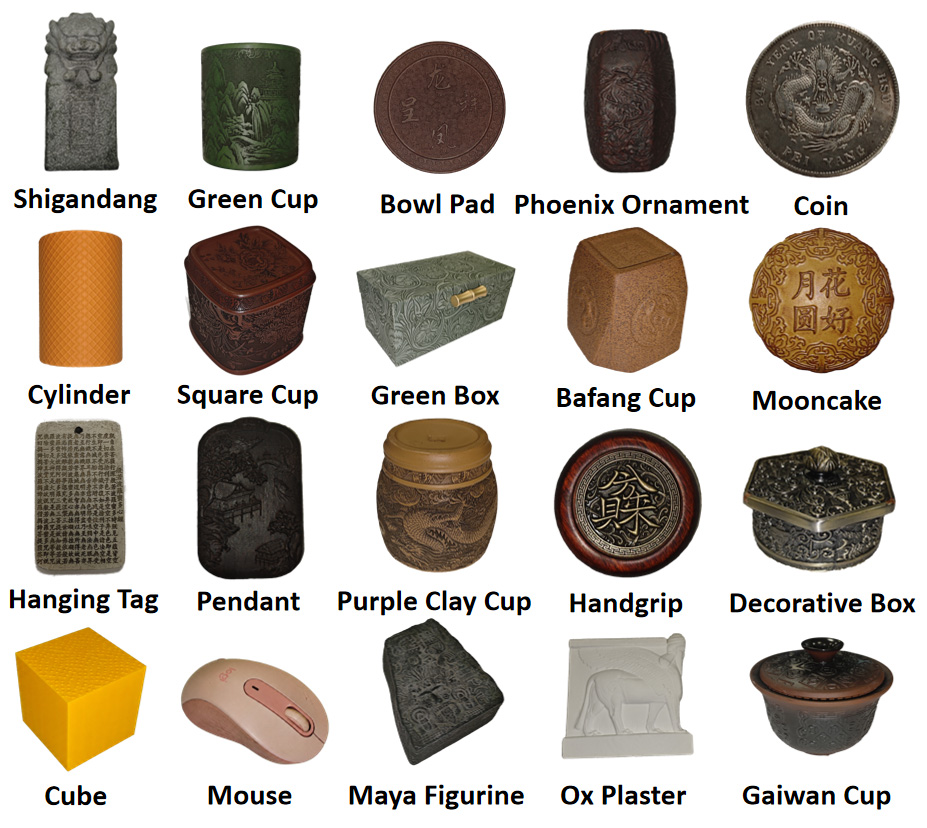}
    \caption{\textbf{Overview of the 20 real objects used in our dataset}. The collection spans flat, curved, cylindrical, and free-form objects with diverse materials and surface relief.}
    \label{fig:dataset_overview}
    \vspace{-1.0\baselineskip}
\end{figure}

\begin{figure}[!b]
    \vspace{-0.5\baselineskip}
    \centering
    \includegraphics[width=\linewidth]{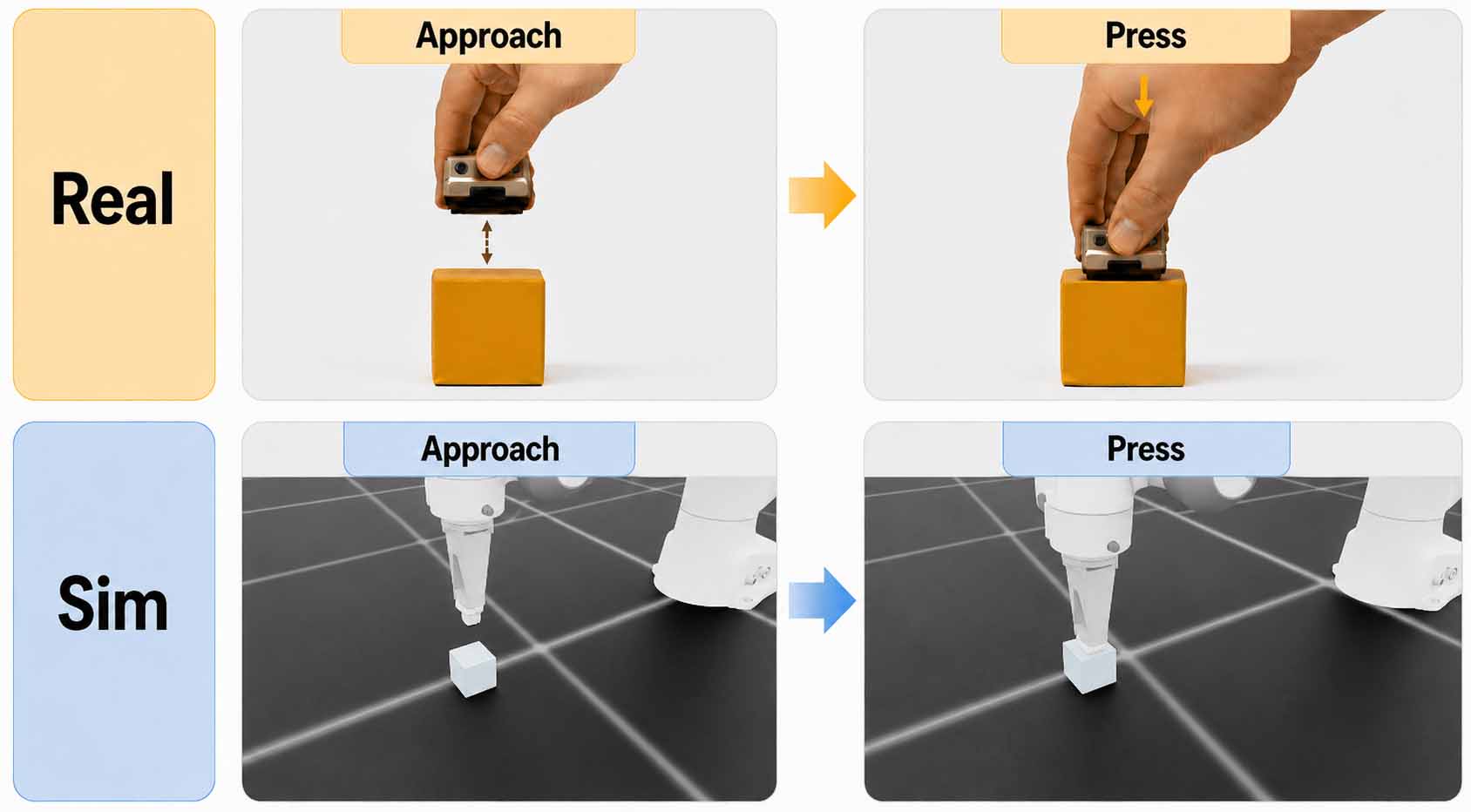}
    \caption{\textbf{Experimental setup for real-to-simulation tactile evaluation}. The real and simulated sensors follow matched approach-and-press procedures.}
    \label{fig:tactile_exp_setup}
\end{figure}

\begin{figure*}[!t]
    \centering
    \includegraphics[width=\textwidth]{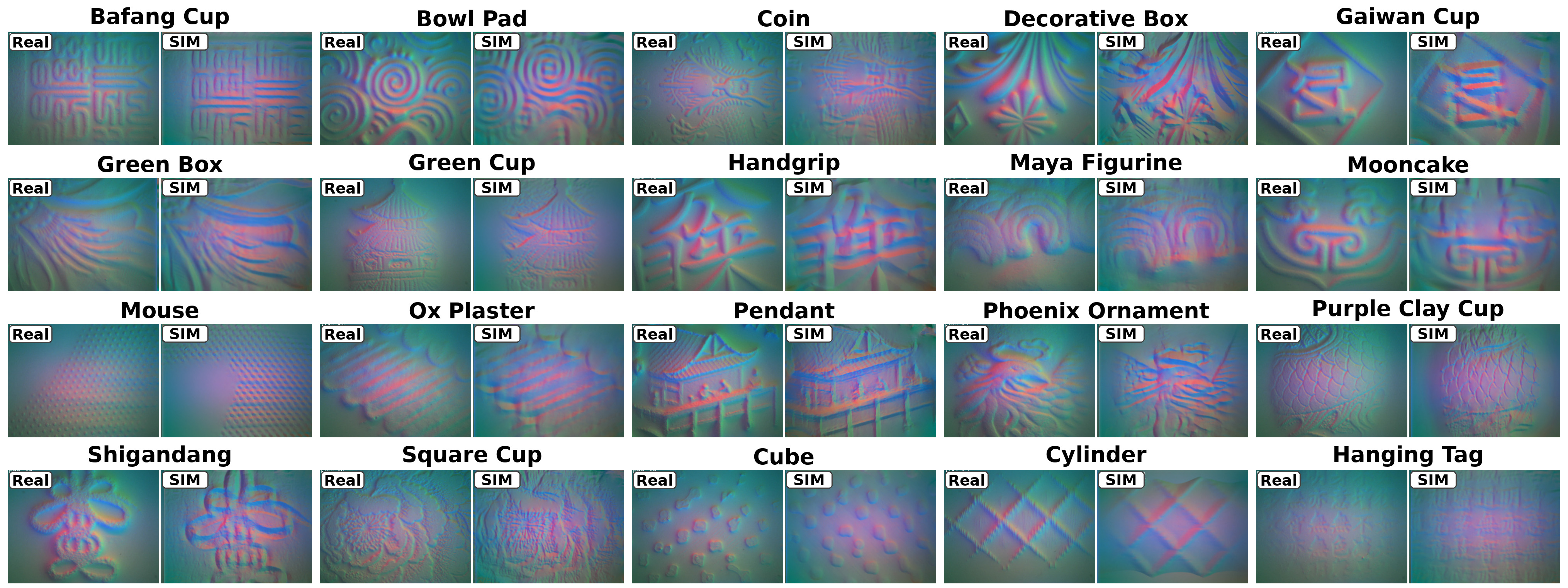}
    \caption{\textbf{Real-to-simulation tactile appearance comparison across 20 objects.} Each object shows a real GelSight Mini measurement and the corresponding simulated rendering under an approximately matched contact region and sensor configuration. Similarity metrics are computed after bounded 2D alignment to compensate for residual image-plane mismatch.}
    \label{fig:tactile_real_sim_compare}
\end{figure*}

\begin{figure*}[!t]
    \centering
    \includegraphics[width=\textwidth]{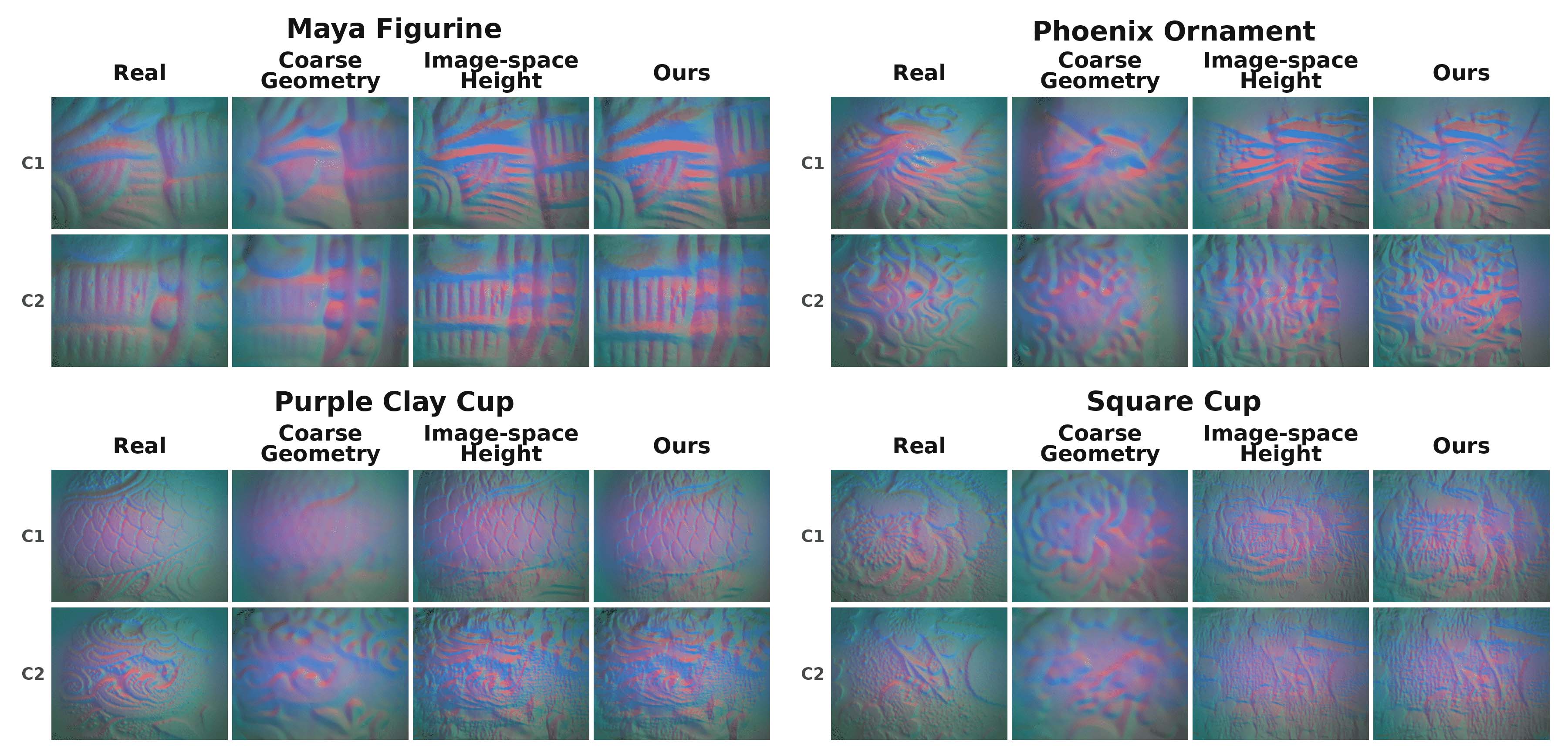}
    \caption{\textbf{Representative tactile-microgeometry ablation on four objects}. Each object is with two contacts (C1–C2). Real images are captured at the corresponding physical contact regions. The evaluation in Table~\ref{tab:tactile_ablation_metrics} covers all 20 objects.}
    \label{fig:normal_ablation}
    \vspace{-0.8\baselineskip}
\end{figure*}

\subsection{TouchTherm Assets Overview}
\label{sec:exp_dataset}
We constructed digital twins for 20 real objects spanning flat, curved,
cylindrical, and free-form geometries, as shown in Fig.~\ref{fig:dataset_overview}.
These objects exhibit smooth, granular, embossed, and anisotropic surface
patterns. As summarized in Fig.~\ref{fig:method_overview}(d), each visuo-tactile
asset comprises a cleaned metric mesh, registered multi-view normal maps, and
reconstructed micro-height fields, which together support simulated tactile
contact and rendering. Excluding object bottoms, planar objects require one
registered RGB/normal view, whereas 3D objects use at least two, typically
5--7, and up to 12 for complex geometries.

For thermal acquisition, synchronized infrared cameras record natural cooling
following heat-gun excitation, following Sec.~\ref{sec:thermal_reconstruction}. Constructing the visuo-tactile assets and acquiring the
thermal data took approximately two to three weeks per modality for the full
collection, corresponding to roughly half a day per object per modality.

\subsection{Tactile Appearance Evaluation and Ablation}
\label{sec:exp_tactile_realism}

First, we evaluate the fidelity of tactile appearance reconstruction across
20 Real--Sim pairs, as shown in Fig.~\ref{fig:tactile_real_sim_compare}. We use two metrics: G-SSIM~\cite{chen2006gradient} for gradient structure and
HF-NCC~\cite{briechle2001template} for high-frequency texture correlation. As this evaluation requires paired Real--Sim data, procedures for data collection are illustrated in Fig.~\ref{fig:tactile_exp_setup}.
Because physical contacts are established manually, we further align each pair
within a fixed search range of $\pm4^{\circ}$ in rotation and
$\pm16$ pixels in translation before evaluating similarity.

In addition to simulation collection, we acquire three repeated physical
GelSight contacts for each object and compute the average pairwise
Real--Real similarity as a reference. Across 20 objects, Real--Real comparisons
yield G-SSIM/HF-NCC of
$0.0954\pm0.0471$/$0.1622\pm0.0917$, while Real--Sim comparisons yield
$0.0712\pm0.0217$/$0.0775\pm0.0575$ (mean $\pm$ standard deviation).

\begin{table}[H]
    \centering
    \refstepcounter{table}
    \label{tab:tactile_ablation_metrics}
    {\raggedright\normalsize\normalfont\noindent TABLE~\thetable:\ \textbf{Tactile ablation over 20 objects, with three contacts evaluated per object.} Results are mean $\pm$ standard deviation: contacts are first averaged within each object, and all objects are weighted. \par}
    \vspace{0.5\baselineskip}
    \scriptsize
    \setlength{\tabcolsep}{5pt}
    \renewcommand{\arraystretch}{1.05}
    \begin{tabular*}{\linewidth}{@{\extracolsep{\fill}}lcc@{}}
        \hline\hline
        \textbf{Method} & \textbf{G-SSIM}$\uparrow$ & \textbf{HF-NCC}$\uparrow$ \\
        \hline
        Coarse Geometry & $0.0480 \pm 0.0283$ & $0.0116 \pm 0.0328$ \\
        Image-space Height & $0.0629 \pm 0.0242$ & $0.0081 \pm 0.0208$ \\
        Ours & $\mathbf{0.0701 \pm 0.0246}$ & $\mathbf{0.0911 \pm 0.0648}$ \\
        \hline\hline
    \end{tabular*}
\end{table}

\textbf{Ablation study.}
\label{sec:tactile_ablation}
We further demonstrate the importance of incorporating fine-grained normal
information and composite surface reconstruction
(Secs.~\ref{sec:normal_estimation_registration}
and~\ref{sec:contact_tactile_rendering}), rather than relying solely on either
coarse surface geometry or direct image-space integration. The comparison
results are shown in Fig.~\ref{fig:normal_ablation} and
Table~\ref{tab:tactile_ablation_metrics}. Specifically, \textit{Coarse
Geometry} uses only the scanned mesh, \textit{Image-space Height} directly integrates camera-frame normal gradients over cropped normal maps, without PnP-based surface projection or tangent-frame normal transformation, and \textit{Ours} combines both sources using
Eq.~(\ref{eq:height_mix}). Our method achieves mean G-SSIM/HF-NCC scores
of 0.0701/0.0911, compared with 0.0480/0.0116 for \textit{Coarse Geometry}
and 0.0629/0.0081 for \textit{Image-space Height}. Consistent with the representative visual comparisons in Fig.~\ref{fig:normal_ablation}, our method achieves the highest object-balanced G-SSIM and HF-NCC across all 20 objects, outperforming both baselines as reported in Table~\ref{tab:tactile_ablation_metrics}.

\begin{figure*}[!t]
    \centering
    \includegraphics[width=\textwidth]{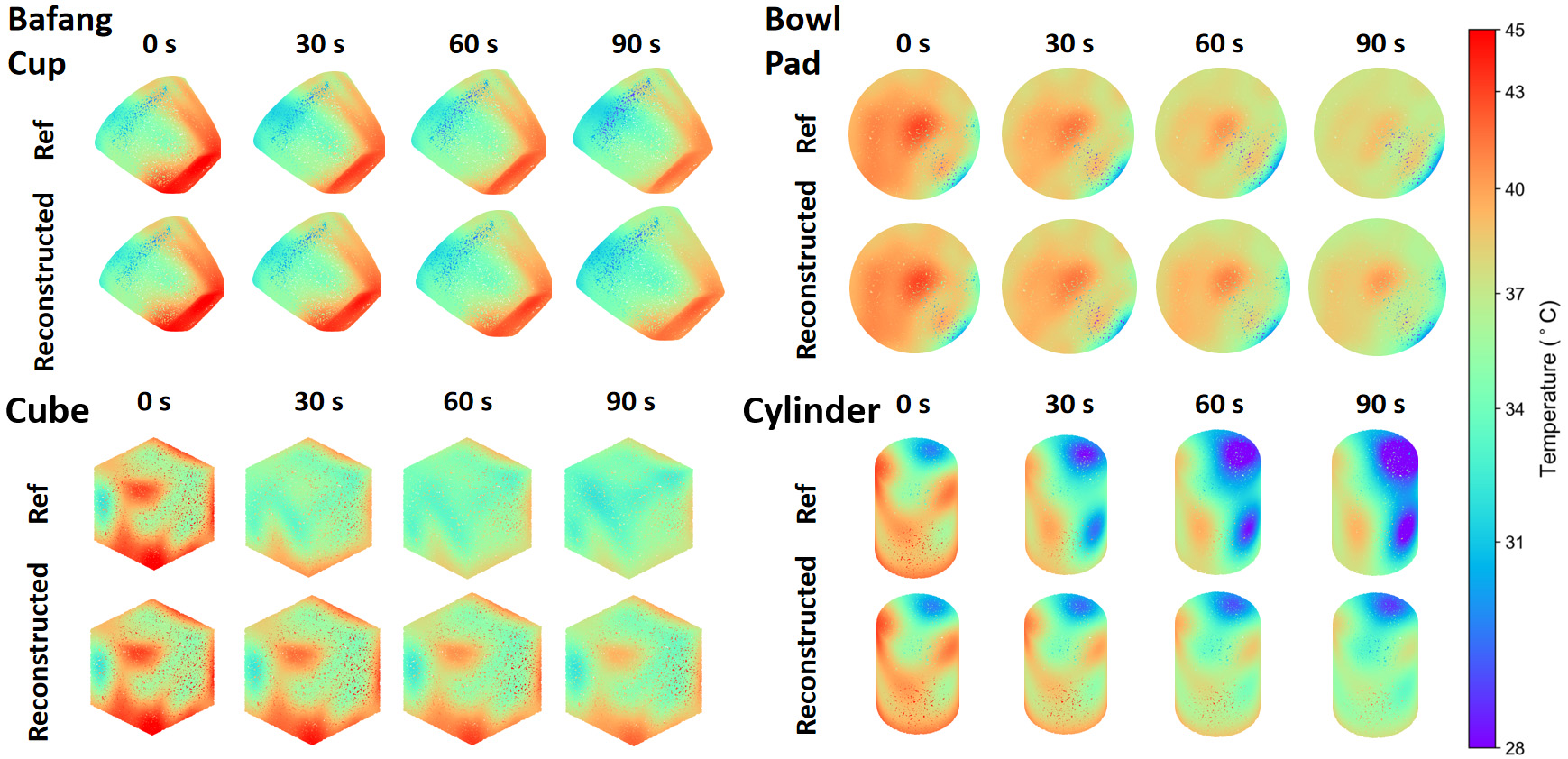}
    \caption{
    \textbf{Comparison of reference and reconstructed thermal evolution
    for four objects at $0$, $30$, $60$, and $90\,\mathrm{s}$.}
    Within each object block, the upper row shows the reference field and the lower row shows the reconstructed field. 
    }
    \label{fig:bafengcup_thermal_result}
\vspace{0.7\baselineskip}
\centering
\refstepcounter{table}
\label{tab:dynamic_thermal_evaluation}
{\raggedright\normalsize\normalfont\noindent TABLE~\thetable:\ {\textbf{Thermal prediction fidelity on test videos.}
On held-out videos, we predict temperatures 30 s and 45 s ahead using only each window's initial temperature field.
Each object's parameters are fitted once from a separate 60-s training capture. We test 20 objects with 15 overlapping windows each, totaling 300 per horizon.
$\Delta\bar{T}_{\mathrm{ref}}$ and $\Delta\bar{T}_{\mathrm{recon}}$ are reference and reconstructed mean surface-temperature changes from each window's start, reported as mean $\pm$ standard deviation.}\par}
\vspace{0.5\baselineskip}
\vspace{0.5\baselineskip}
\resizebox{\textwidth}{!}{%
\begin{tabular}{llccccc}
\hline\hline
\textbf{Method}
& \textbf{Time}
& $\boldsymbol{\Delta\bar{T}_{\mathrm{ref}}}$ \textbf{($^{\circ}\mathrm{C}$)}
& $\boldsymbol{\Delta\bar{T}_{\mathrm{recon}}}$ \textbf{($^{\circ}\mathrm{C}$)}
& \textbf{Surface MAE ($^{\circ}\mathrm{C}$)} $\boldsymbol{\downarrow}$
& \textbf{Hotspot F1} $\boldsymbol{\uparrow}$
& \textbf{Gradient RMSE ($^{\circ}\mathrm{C}$)} $\boldsymbol{\downarrow}$ \\
\hline
Full ($\alpha+h$)
& $30\,\mathrm{s}$
& $-0.765 \pm 0.443$
& $\mathbf{-0.706 \pm 0.317}$
& $\mathbf{0.465 \pm 0.240}$
& $\mathbf{0.793 \pm 0.179}$
& $\mathbf{0.0079 \pm 0.0036}$ \\
Full ($\alpha+h$)
& $45\,\mathrm{s}$
& $-1.033 \pm 0.520$
& $\mathbf{-1.043 \pm 0.462}$
& $\mathbf{0.592 \pm 0.266}$
& $\mathbf{0.708 \pm 0.246}$
& $\mathbf{0.0097 \pm 0.0039}$ \\
\hline
No $\alpha$
& $30\,\mathrm{s}$
& $-0.765 \pm 0.443$
& $-0.674 \pm 0.208$
& $0.470 \pm 0.263$
& $0.783 \pm 0.182$
& $0.0081 \pm 0.0040$ \\
No $\alpha$
& $45\,\mathrm{s}$
& $-1.033 \pm 0.520$
& $-0.997 \pm 0.304$
& $\mathbf{0.592 \pm 0.279}$
& $0.698 \pm 0.230$
& $0.0100 \pm 0.0045$ \\
\hline
No $h$
& $30\,\mathrm{s}$
& $-0.765 \pm 0.443$
& $0.001 \pm 0.017$
& $0.820 \pm 0.411$
& $0.705 \pm 0.165$
& $0.0083 \pm 0.0039$ \\
No $h$
& $45\,\mathrm{s}$
& $-1.033 \pm 0.520$
& $0.001 \pm 0.025$
& $1.083 \pm 0.493$
& $0.632 \pm 0.187$
& $0.0103 \pm 0.0043$ \\
\hline\hline
\end{tabular}%
}
\vspace{-0.6\baselineskip}
\end{figure*}

\subsection{Dynamic Thermal Evaluation}
\label{sec:exp_thermal}

As shown in Fig.~\ref{fig:bafengcup_thermal_result}, the reconstructed temperature
fields closely match the reference fields over time. This demonstrates that the
model captures both global cooling trends and spatial temperature patterns
over long prediction horizons. To assess the role of the identified parameters
(Sec.~\ref{sec:physics_reg}), we compare the full model ($\alpha+h$) with two
ablations, no-$\alpha$ ($\alpha=0$) and no-$h$ ($h=0$), on all 20 objects at
30\,s and 45\,s prediction horizons. We report mean surface-temperature change
and surface MAE for global cooling and vertex-wise accuracy, respectively, and
Gradient RMSE for local temperature variation. Because surface MAE does not
capture the spatial agreement of high-temperature regions, we additionally
report Hotspot F1, where hotspots are defined independently in each field as
$T \geq P_1+0.9(P_{99}-P_1)$, with $P_1$ and $P_{99}$ denoting the first and
99th temperature percentiles.

As shown in Table~\ref{tab:dynamic_thermal_evaluation}, the full model closely
matches the reference cooling at both prediction horizons. Removing $\alpha$
causes only minor degradation, whereas removing $h$ largely suppresses cooling
and substantially increases the surface MAE. These results indicate that
ambient heat exchange primarily governs global cooling, while surface diffusion
provides smaller improvements in local temperature prediction.

\subsection{Downstream Applications}
\label{sec:exp_tactile_recognition}

We demonstrate the utility of the proposed assets in downstream tasks.
Due to space constraints, we focus on demonstrations that
highlight their potential for broader use, rather than providing a comprehensive
evaluation.

\noindent\textbf{Synthetic-to-Real Tactile Object Recognition.}
We evaluate whether the reconstructed tactile microgeometry provides
transferable object-discriminative cues for synthetic-to-real object
recognition. We compare two settings, Coarse Geometry and TouchTherm, using
an identical ImageNet-pretrained ResNet-18 trained only on simulated
GelSight images. For each object, 90 simulated contacts are used for
training and 10 additional contacts for validation. Both settings use the
same contact configurations, data split, training schedule, augmentations,
and six random seeds, making the tactile representation the only variable.
For real-world evaluation, we use 100 real GelSight images (5 per object across 20 objects) to perform a global sensor-level color calibration for TacEx. The resulting renderer parameters are shared across all objects, while the remaining 20 images per object are held out for evaluation. No real image is used for classifier training or validation.

We report Top-1 and Top-3 accuracy and Macro-F1, defined as the unweighted mean of the per-class F1 scores.
Across six random seeds, Coarse Geometry achieves Top-1, Top-3, and
Macro-F1 scores of $20.0\pm4.4\%$, $35.0\pm2.3\%$, and
$13.4\pm3.7\%$, respectively. In contrast, TouchTherm achieves
$34.0\pm2.7\%$, $63.0\pm1.7\%$, and $29.3\pm2.5\%$, respectively.
The improvement is consistent across all three metrics, with TouchTherm
increasing Top-1 accuracy by 14.0 percentage points and nearly doubling
both Top-3 accuracy and Macro-F1. These results indicate that the registered tactile microgeometry provides object-dependent surface cues that improve synthetic-to-real tactile recognition over coarse collision geometry alone.

\begin{figure}[t]
    \centering
    \includegraphics[width=\linewidth]{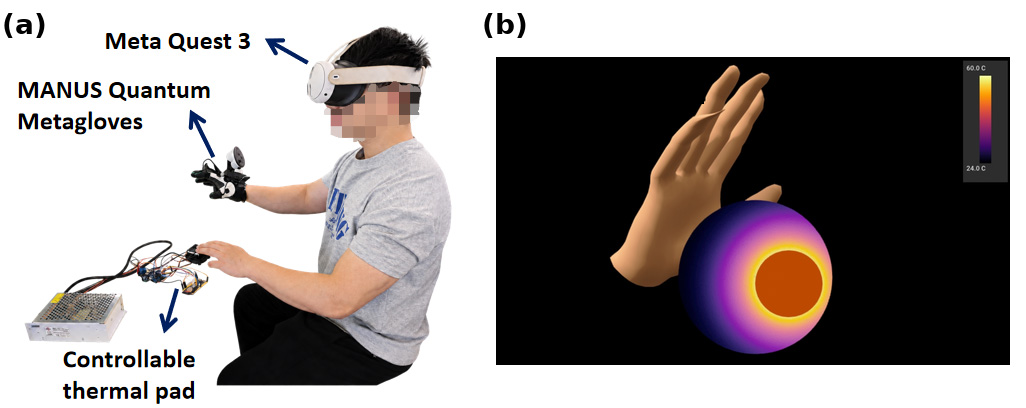}
    \caption{\textbf{Thermal-feedback VR system}. (a) VR and thermal-feedback hardware and (b) interaction with a reconstructed dynamic thermal field.}
    \label{fig:downstream_result}
    \vspace{-1.0\baselineskip}
\end{figure}

\noindent\textbf{Thermal Feedback in VR}.
We further evaluate the usability of the reconstructed thermal properties.
Ten consenting, self-reported healthy participants (8 male and 2 female, aged 18–28) used the wearable platform in
Fig.~\ref{fig:downstream_result} to explore cold, warm, hot, and continuously
cooling regions. On a seven-point scale (with 7 denoting the highest rating),
visual--thermal consistency, cooling consistency, response timeliness, realism,
and comfort received scores of
$6.30\pm0.95$, $6.20\pm1.62$, $5.40\pm1.07$,
$6.10\pm0.99$, and $6.40\pm1.26$, respectively.
Overall, all participants recognized the gradual cold--warm--hot changes across different object regions, with 7 participants rating the rendered cues as highly clear (the highest rating).
All ten participants reported no discomfort, indicating that the system can provide intuitive and comfortable thermal feedback.

\section{Conclusion}

We presented \textbf{TouchTherm}, a framework for constructing
simulation-ready multimodal digital twins that integrate coarse visual
geometry, tactile microgeometry, and a dynamic thermal field.
By separating tactile micro-height fields from collision geometry, the
framework preserves contact-scale surface details without increasing
collision complexity. It also reconstructs a dynamic thermal field from
multiview infrared observations, enabling spatiotemporal temperature queries.
Experiments on 20 objects demonstrate improved tactile rendering compared
with coarse-geometry and image-space baselines, as well as accurate thermal
predictions on held-out data, with surface MAEs of
$0.465\,^{\circ}\mathrm{C}$ and $0.592\,^{\circ}\mathrm{C}$ at 30\,s and
45\,s, respectively. In synthetic-to-real tactile object recognition, TouchTherm
consistently outperforms coarse geometry across Top-1, Top-3, and Macro-F1.
An application using a wearable glove further
demonstrates spatially and temporally varying thermal feedback in VR.

\textbf{Limitations and future work.}
The current framework uses microgeometry only for tactile rendering and does
not model its effects on contact mechanics. Its dynamic thermal field focuses
on natural cooling and does not capture bidirectional heat transfer during human
or robotic contact. Future work will integrate microgeometry with compliant
contact and friction models and extend the dynamic thermal field to support
contact-dependent, bidirectional heat transfer.

\bibliography{egbib}
\bibliographystyle{IEEEtran}

\end{document}